\documentclass[11pt]{article}

\usepackage{acl2016}
\usepackage{times}
\usepackage{url}
\usepackage{latexsym}

\aclfinalcopy 

\usepackage{amsfonts}
\usepackage{amsmath}
\usepackage{booktabs}
\usepackage{graphicx}
\usepackage{paralist}
\usepackage{tabularx}
\usepackage[usenames,dvipsnames]{xcolor}
\usepackage{enumitem}

\title{Siamese CBOW: Optimizing Word Embeddings\\ for Sentence Representations}

\author{
\begin{tabular}{ccc}
\textbf{Tom Kenter}$^1$ &
\textbf{Alexey Borisov}$^{1,\, 2}$ &
\textbf{Maarten de Rijke}$^1$ \\
{\tt tom.kenter@uva.nl} &
  {\tt alborisov@yandex-team.ru} &
  {\tt derijke@uva.nl}
\end{tabular}\\[2ex]
$^1$~University of Amsterdam,  Amsterdam \\
$^2$~Yandex, Moscow
}

\begin{document}

\maketitle

\begin{abstract}
We present the \emph{Siamese Continuous Bag of Words} (Siamese CBOW) model, a neural network for efficient estimation of high-quality sentence embeddings.
Averaging the embeddings of words in a sentence has proven to be a surprisingly successful and efficient way of obtaining sentence embeddings.
However, word embeddings trained with the methods currently available are not optimized for the task of sentence representation, and, thus, likely to be suboptimal.
Siamese CBOW handles this problem by training word embeddings directly for the purpose of being averaged.
The underlying neural network learns word embeddings by predicting, from a sentence representation, its surrounding sentences.
We show the robustness of the Siamese CBOW model by evaluating it on 20 datasets stemming from a wide variety of sources.
%
\end{abstract}


\section{Introduction}\label{sec:introduction}

Word embeddings have proven to be beneficial in a variety of tasks in NLP such as machine translation \cite{zou2013bilingualwe}, parsing~\cite{chen2014fast}, semantic search~\cite{reinanda2015entityaspects,voskarides-learning-2015}, and tracking the meaning of words and concepts over time \cite{kim2014temporal,kenter-ad-hoc-2015}.
It is not evident, however, how word embeddings should be combined to represent larger pieces of text, like sentences, paragraphs or documents. Surprisingly, simply averaging word embeddings of all words in a text has proven to be a strong baseline or feature across a multitude of tasks \cite{faruqui2014retrofitting,yu2014answerselection,gershman2015phrase,kenter-short-2015}.

Word embeddings, however, are not optimized specifically for representing sentences.
In this paper we present a model for obtaining word embeddings that are tailored specifically for the task of averaging them.
We do this by directly including a comparison of sentence embeddings---the averaged embeddings of the words they contain---in the cost function of our network.

Word embeddings are typically trained in a fast and scalable way from unlabeled training data. As the training data is unlabeled, word embeddings are usually not task-specific. Rather, word embeddings trained on a large training corpus, like the ones from \cite{collobert2008unified,mikolov_distributed_2013} are employed across different tasks \cite{socher2012semantic,kenter-short-2015,hu2014convolutional}.
These two qualities---%
\begin{inparaenum}[(i)]
    \item being trainable from large quantities of unlabeled data in a reasonable amount of time, and 
    \item  robust performance across different tasks---%
\end{inparaenum}%
are highly desirable and allow word embeddings to be used in many large-scale applications. In this work we aim to optimize word embeddings for sentence representations in the same manner. We want to produce general purpose sentence embeddings that should score robustly across multiple test sets, and we want to leverage large amounts of unlabeled training material.

In the word2vec algorithm, \newcite{mikolov_efficient_2013} construe a supervised training criterion for obtaining word embeddings from unsupervised data, by predicting, for every word, its surrounding words. We apply this strategy at the sentence level, where we aim to predict a sentence from its adjacent sentences \cite{kiros2015skipthought,hill2016learning}. This allows us to use unlabeled training data, which is easy to obtain; the only restriction is that documents need to be split into sentences and that the order between sentences is preserved.

The main research question we address is whether directly optimizing word embeddings for the task of being averaged to produce sentence embeddings leads to word embeddings that are better suited for this task than word2vec does.
Therefore, we test the embeddings in an unsupervised learning scenario. We use 20 evaluation sets that stem from a wide variety of sources (newswire, video descriptions, dictionary descriptions, microblog posts).
Furthermore, we analyze the time complexity of our method and compare it to our baselines methods. 

Summarizing, our main contributions are:
\begin{itemize}[leftmargin=0.35cm,nosep]
\item We present Siamese CBOW, an efficient neural network architecture for obtaining high-quality word embeddings, directly optimized for sentence representations;
\item We evaluate the embeddings produced by Siamese CBOW on 20 datasets, originating from a range of sources (newswire, tweets, video descriptions), and demonstrate the robustness of embeddings across different settings.
\end{itemize}



\section{Siamese CBOW}\label{sec:our_approach}

\begin{figure*}[ht]
  \centering
  \includegraphics[width=\textwidth]{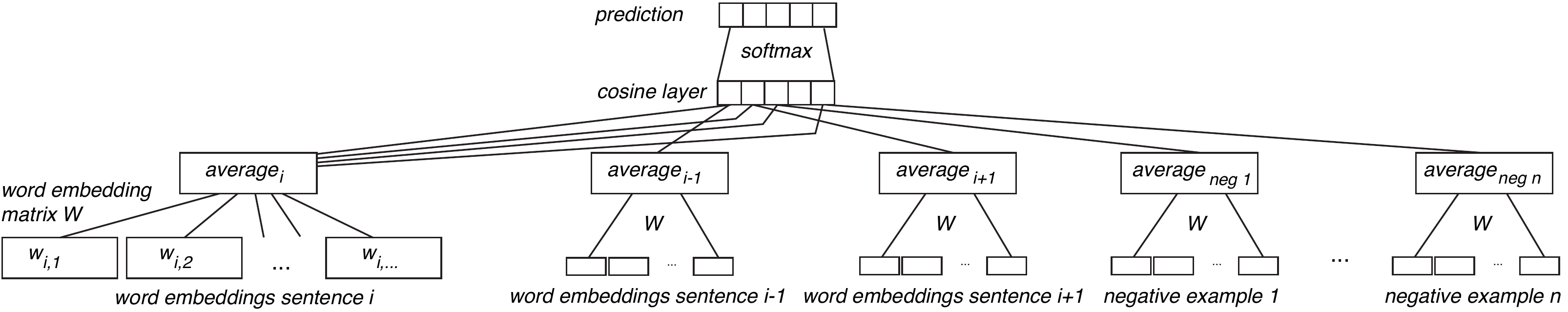}
  \caption{Siamese CBOW network architecture. (Input projection layer omitted.)}
  \label{figure:ssnArchitecture}
\end{figure*}
We present the \emph{Siamese Continuous Bag of Words} (CBOW) model, a neural network for efficient estimation of high-quality sentence embeddings.
Quality should manifest itself in embeddings of semantically close sentences being similar to one another, and embeddings of semantically different sentences being dissimilar.
%
%
%
An efficient and surprisingly successful way of computing a sentence embedding is to average the embeddings of its constituent words.
Recent work uses pre-trained word embeddings (such as word2vec and GloVe) for this task, which are not optimized for sentence representations.
Following these approaches, we compute sentence embeddings by averaging word embeddings, but we optimize word embeddings directly for the purpose of being averaged.


\subsection{Training objective}\label{sec:our_approach:training_objective}

We construct a supervised training criterion by having our network predict sentences occurring next to each other in the training data.
Specifically, for a pair of sentences $(s_i, s_j)$, we define a probability $p(s_i, s_j)$ that reflects how likely it is for the sentences to be adjacent to one another in the training data.
We compute the probability $p(s_i, s_j)$ using a softmax function:
\begin{equation}\label{equation:probability}
  p_{\theta}(s_i, s_j) = \frac{e^{\cos(\mathbf{s_i^{\theta}}, \mathbf{s_j^{\theta}})}}{\sum_{s_{'} \in S} e ^ {\cos(\mathbf{s_i^{\theta}}, \mathbf{s_{'}^{\theta}})}} ,
\end{equation}
where $\mathbf{s}_x^{\theta}$ denotes the embedding of sentence $s_x$, based on the model parameters $\theta$.
In theory, the summation in the denominator of Equation~\ref{equation:probability} should range over all possible sentences~$S$, which is not feasible in practice.
Therefore, we replace the set $S$ with the union of the set $S^+$ of sentences that occur next to the sentence $s_i$ in the training data, and $S^-$, a set of $n$ randomly chosen sentences that are not observed next to the sentence $s_i$ in the training data.
The loss function of the network is categorical cross-entropy:
\vspace*{-.2\baselineskip}
\begin{eqnarray*}
  L &=& - \sum_{s_j \in \{ S^+ \thinspace\cup\thinspace S^- \}} p(s_i, s_j) \cdot \log( p_{\theta}(s_i, s_j )),
\end{eqnarray*}

\vspace*{-.2\baselineskip}\noindent%
where $p(\cdot)$ is the target probability the network should produce, and $p_{\theta}(\cdot)$ is the prediction it estimates based on parameters $\theta$, using Equation~\ref{equation:probability}.
The target distribution simply is:
\[
p(s_i, s_j) = \left\{
  \begin{array}{ll}
    \frac{1}{|S^+|}, & \text{if } s_j \in S^+ \\
    0, & \text{if } s_j \in S^-.
  \end{array}
\right.
\]
I.e., if there are 2 positive examples (the sentences preceding and following the input sentence) and 2 negative examples, the target distribution is (0.5,~0.5,~0,~0).

\subsection{Network architecture}\label{sec:our_approach:network_architecture}

Figure~\ref{figure:ssnArchitecture} shows the architecture of the proposed Siamese CBOW network.
The input is a projection layer that selects embeddings from a word embedding matrix $W$ (that is shared across inputs) for a given input sentence.
The word embeddings are averaged in the next layer, which yields a sentence representation with the same dimensionality as the input word embeddings (the boxes labeled $\textit{average}_i$ in Figure~\ref{figure:ssnArchitecture}).
The cosine similarities between the sentence representation for $\textit{sentence}_i$ and the other sentences are calculated in the penultimate layer and a softmax is applied in the last layer to produce the final probability distribution.

\subsection{Training}\label{sec:our_approach:training}

The weights in the word embedding matrix are the only trainable parameters in the Siamese CBOW network.
They are updated using stochastic gradient descent.
The initial learning rate is monotonically decreased proportionally to the number of training batches.



\section{Experimental Setup}\label{sec:experimental setup}

To test the efficacy of our siamese network for producing sentence embeddings we use multiple test sets.
We use Siamese CBOW to learn word embeddings from an unlabeled corpus.
For every sentence pair in the test sets, we compute two sentence representations by averaging the word embeddings of each sentence.
Words that are missing from the vocabulary and, hence, have no word embedding, are omitted.
The cosine similarity between the two sentence vectors is produced as a final semantic similarity score.

As we want a clean way to directly evaluate the embeddings on multiple sets we train our model and the models we compare with on exactly the same training data.
We do not compute extra features, perform extra preprocessing steps or incorporate the embeddings in supervised training schemes.
Additional steps like these are very likely to improve evaluation scores, but they would obscure our main evaluation purpose in this paper, which is to directly test the embeddings. 


\subsection{Data}

We use the Toronto Book Corpus\footnote{The corpus can be downloaded from \url{http://www.cs.toronto.edu/~mbweb/}; cf.\ \cite{torontoBookCorpus}.} to train word embeddings.
This corpus contains 74,004,228 already pre-processed sentences in total, which are made up of 1,057,070,918 tokens, originating from 7,087 unique books.
In our experiments, we consider tokens appearing 5 times or more, which leads to a vocabulary of 315,643 words.

\subsection{Baselines}

We employ two baselines for producing sentence embeddings in our experiments. 
We obtain similarity scores between sentence pairs from the baselines in the same way as the ones produced by Siamese CBOW, i.e., we calculate the cosine similarity between the sentence embeddings they produce.

\paragraph{Word2vec} We average word embeddings trained with word2vec.\footnote{The code is available from {\small\url{https://code.google.com/archive/p/word2vec/}}.}
We use both architectures, Skipgram and CBOW, and apply default settings: minimum word frequency 5, word embedding size 300, context window 5, sample threshold $\text{10}^{\text{-5}}$, no hierarchical softmax, 5 negative examples.

\paragraph{Skip-thought} As a second baseline we use the sentence representations produced by the skip-thought architecture \cite{kiros2015skipthought}.\footnote{The code and the trained models can be downloaded from \url{https://github.com/ryankiros/skip-thoughts/}.} Skip-thought is a recently proposed method that learns sentence representations in a different way from ours, by using recurrent neural networks. This allows it to take word order into account. As it trains sentence embeddings from unlabeled data, like we do, it is a natural baseline to consider.

\medskip\noindent%
Both methods are trained on the Toronto Book Corpus, the same corpus used to train Siamese CBOW.
We should note that as we use skip-thought vectors as trained by \newcite{kiros2015skipthought}, skip-thought has an advantage over both word2vec and Siamese CBOW as the vocabulary used for encoding sentences contains 930,913 words, three times the size of the vocabulary that we use.

\begin{table*}[ht]
  \centering
  \caption{Results on SemEval datasets in terms of Pearson's $r$ (Spearman's $r$). Highest scores, in terms of Pearson's $r$, are displayed in bold. Siamese CBOW runs statistically significantly different from the word2vec CBOW baseline runs are marked with a $\dagger$. See \S\ref{sec:evaluation} for a discussion of the statistical test used.}
  \begin{tabular}{p{.15\textwidth} >{\centering\arraybackslash}p{.16\textwidth} >{\centering\arraybackslash}p{.16\textwidth} >{\centering\arraybackslash}p{.16\textwidth} >{\centering\arraybackslash}p{.16\textwidth}}
  \toprule
  Dataset &  w2v skipgram & w2v CBOW & skip-thought & Siamese CBOW \\
  \midrule
  2012 & & & & \\
  \midrule
  MSRpar &  $.3740~(.3991$) & $.3419~(.3521$) & $.0560~(.0843$) & $\mathbf{.4379}^\dagger~(.4311$) \\
  MSRvid &  $.5213~(.5519$) & $.5099~(.5450$) & $\mathbf{.5807}~(.5829$) & $.4522^\dagger~(.4759$) \\
  OnWN &  $.6040~(.6476$) & $.6320~(.6440$) & $.6045~(.6431$) & $\mathbf{.6444}^\dagger~(.6475$) \\
  SMTeuroparl &  $.3071~(.5238$) & $.3976~(.5310$) & $.4203~(.4999$) & $\mathbf{.4503}^\dagger~(.5449$) \\
  SMTnews &  $\mathbf{.4487}~(.3617$) & $.4462~(.3901$) & $.3911~(.3628$) & $.3902^\dagger~(.4153$) \\
  \midrule
  2013 & & & & \\
  \midrule
  FNWN &  $\mathbf{.3480}~(.3401$) & $.2736~(.2867$) & $.3124~(.3511$) & $.2322^\dagger~(.2235$) \\
  OnWN &  $.4745~(.5509$) & $\mathbf{.5165}~(.6008$) & $.2418~(.2766$) & $.4985^\dagger~(.5227$) \\
  SMT &  $.1838~(.2843$) & $.2494~(.2919$) & $\mathbf{.3378}~(.3498)$ & $.3312^\dagger~(.3356$) \\
  headlines &  $.5935~(.6044$) & $.5730~(.5766$) & $.3861~(.3909$) & $\mathbf{.6534}^\dagger~(.6516$) \\
  \midrule
  2014 & & & & \\
  \midrule
  OnWN &  $.5848~(.6676$) & $.6068~(.6887$) & $.4682~(.5161$) & $\mathbf{.6073}^\dagger~(.6554$) \\
  deft-forum &  $.3193~(.3810$) & $.3339~(.3507$) & $.3736~(.3737$) & $\mathbf{.4082}^\dagger~(.4188$) \\
  deft-news &  $.5906~(.5678$) & $.5737~(.5577$) & $.4617~(.4762$) & $\mathbf{.5913}^\dagger~(.5754$) \\
  headlines &  $.5790~(.5544$) & $.5455~(.5095$) & $.4031~(.3910$) & $\mathbf{.6364}^\dagger~(.6260$) \\
  images &  $.5131~(.5288$) & $.5056~(.5213$) & $.4257~(.4233$) & $\mathbf{.6497}^\dagger~(.6484$) \\
  tweet-news &  $.6336~(.6544$) & $.6897~(.6615$) & $.5138~(.5297$) & $\mathbf{.7315}^\dagger~(.7128$) \\
  \midrule
  2015 & & & & \\
  \midrule
  answ-forums &  $.1892~(.1463$) & $.1767~(.1294$) & $\mathbf{.2784}~(.1909$) & $.2181^{~}~(.1469$) \\
  answ-students &  $.3233~(.2654$) & $.3344~(.2742$) & $.2661~(.2068$) & $\mathbf{.3671}^\dagger~(.2824$) \\
  belief &  $.2435~(.2635$) & $.3277~(.3280$) & $.4584~(.3368$) & $\mathbf{.4769}^{~}~(.3184$) \\
  headlines &  $.1875~(.0754$) & $.1806~(.0765$) & $.1248~(.0464$) & $\mathbf{.2151}^\dagger~(.0846$) \\
  images &  $.2454~(.1611$) & $.2292~(.1438$) & $.2100~(.1220$) & $\mathbf{.2560}^\dagger~(.1467$) \\
  \bottomrule
 \end{tabular}
 \label{table:resultsToBoCo}
\end{table*}

\subsection{Evaluation}\label{sec:evaluation}

We use 20 SemEval datasets from the SemEval semantic textual similarity task in 2012, 2013, 2014 and 2015 \cite{agirre_semeval-2012_2012,agirre2013sem,agirre2014semeval2014,agirre2015semeval2015t2}, which consist of sentence pairs from a wide array of sources (e.g., newswire, tweets, video descriptions) that have been manually annotated by multiple human assessors on a 5 point scale (1: semantically unrelated, 5: semantically similar).
In the ground truth, the final similarity score for every sentence pair is the mean of the annotator judgements, and as such can be a floating point number like 2.685.

The evaluation metric used by SemEval, and hence by us, is Pearson's $r$.
As Spearman's $r$ is often reported as well, we do so too.

\paragraph{Statistical significance} To see whether Siamese CBOW yields significantly different scores for the same input sentence pairs from word2vec CBOW---the method it is theoretically most similar to---we compute Wilcoxon signed-rank test statistics between all runs on all evaluation sets.
Runs are considered statistically significantly different for p-values~$<$~0.0001.

\subsection{Network}\label{sec:network details}

To comply with results reported in other research \cite{mikolov_distributed_2013,kusner2015word} we fix the embedding size to 300 and only consider words appearing 5 times or more in the training corpus.
We use 2 negative examples (see \S\ref{sec:nrNegExamples} for an analysis of different settings).
The embeddings are initialized randomly, by drawing from a normal distribution with $\mu$ = 0.0 and $\sigma$ = 0.01. 
The batch size is 100. The initial learning rate $\alpha$ is 0.0001, which we obtain by observing the loss on the training data.
Training consists of one epoch.

We use Theano \cite{theano2016} to implement our network.\footnote{The code for Siamese CBOW is available under an open-source license at \url{https://bitbucket.org/TomKenter/siamese-cbow}.}
We ran our experiments on GPUs in the DAS5 cluster \cite{bal2016medium}.


\section{Results}\label{sec:results}

In this section we present the results of our experiments, and analyze the stability of Siamese CBOW with respect to its (hyper)parameters. 

\subsection{Main experiments}\label{sec:main experiments}

In Table~\ref{table:resultsToBoCo}, the results of Siamese CBOW on 20 SemEval datasets are displayed, together with the results of the baseline systems.
As we can see from the table, Siamese CBOW outperforms the baselines in the majority of cases (14 out of 20).
The very low scores of skip-thought on MSRpar appear to be a glitch, which we will ignore.

It is interesting to see that for the set with the highest average sentence length (2013 SMT, with 24.7 words per sentence on average) Siamese CBOW is very close to skip-thought, the best performing baseline.
In terms of lexical term overlap, unsurprisingly, all methods have trouble with the sets with little overlap (2013 FNWN, 2015 answers-forums, which both have 7\% lexical overlap).
It is interesting to see, however, that for the next two sets (2015 belief and 2012 MSRpar, 11\% and 14\% overlap respectively) Siamese CBOW manages to get the best performance.
The highest performance on all sets is 0.7315 Pearson's $r$ of Siamese CBOW on the 2014 tweet-news set.
This figure is not very far from the best performing SemEval run that year which has 0.792 Pearson's $r$.
This is remarkable as Siamese CBOW is completely unsupervised, while the NTNU system which scored best on this set \cite{lynum2014ntnums} was optimized using multiple training sets.

In recent work, \newcite{hill2016learning} present FastSent, a model similar to ours (see \S\ref{sec:related work} for a more elaborate discussion); results are not reported for all evaluation sets we use, and hence, we compare the results of FastSent and Siamese CBOW separately, in Table~\ref{table:fastSent}.

\begin{table}[h]
  \centering
  \caption{Results on SemEval 2014 datasets in terms of Pearson's $r$ (Spearman's $r$). Highest scores (in Pearson's $r$) are displayed in bold. FastSent results are reprinted from \protect\cite{hill2016learning} where they are reported in two-digit precision.}
  \begin{tabular}{p{.11\textwidth} >{\centering\arraybackslash}p{.13\textwidth} >{\centering\arraybackslash}p{.16\textwidth} }
  \toprule
  Dataset & FastSent & Siamese CBOW \\
  \midrule
  OnWN &       $\mathbf{.74}~(.70)$ & $.6073~(.6554$) \\
  deft-forum & $\mathbf{.41}~(.36)$ & $.4082~(.4188$) \\
  deft-news &  $.58~(.59)$          & $\mathbf{.5913}~(.5754$) \\
  headlines &  $.57~(.59)$          & $\mathbf{.6364}~(.6260$) \\
  images &     $\mathbf{.74}~(.78)$ & $.6497~(.6484$) \\
  tweet-news & $.63~(.66)$          & $\mathbf{.7315}~(.7128$) \\
  \bottomrule
 \end{tabular}
 \label{table:fastSent}
\end{table}

FastSent and Siamese CBOW each outperform the other on half of the evaluation sets, which clearly suggests that the differences between the two methods are complementary.\footnote{The comparison is to be interpreted with caution as it is not evident what vocabulary was used for the experiments in \cite{hill2016learning}; hence, the differences observed here might simply be due to differences in vocabulary coverage.}

\subsection{Analysis}\label{sec:analysis}

Next, we investigate the stability of Siamese CBOW with respect to its hyper-parameters.
In particular, we look into stability across iterations, different numbers of negative examples, and the dimensionality of the embeddings.   
Other parameter settings are set as reported in \S\ref{sec:network details}.

\subsubsection{Performance across iterations}

\begin{figure*}[ht]
  \centering
  \includegraphics[width=\textwidth]{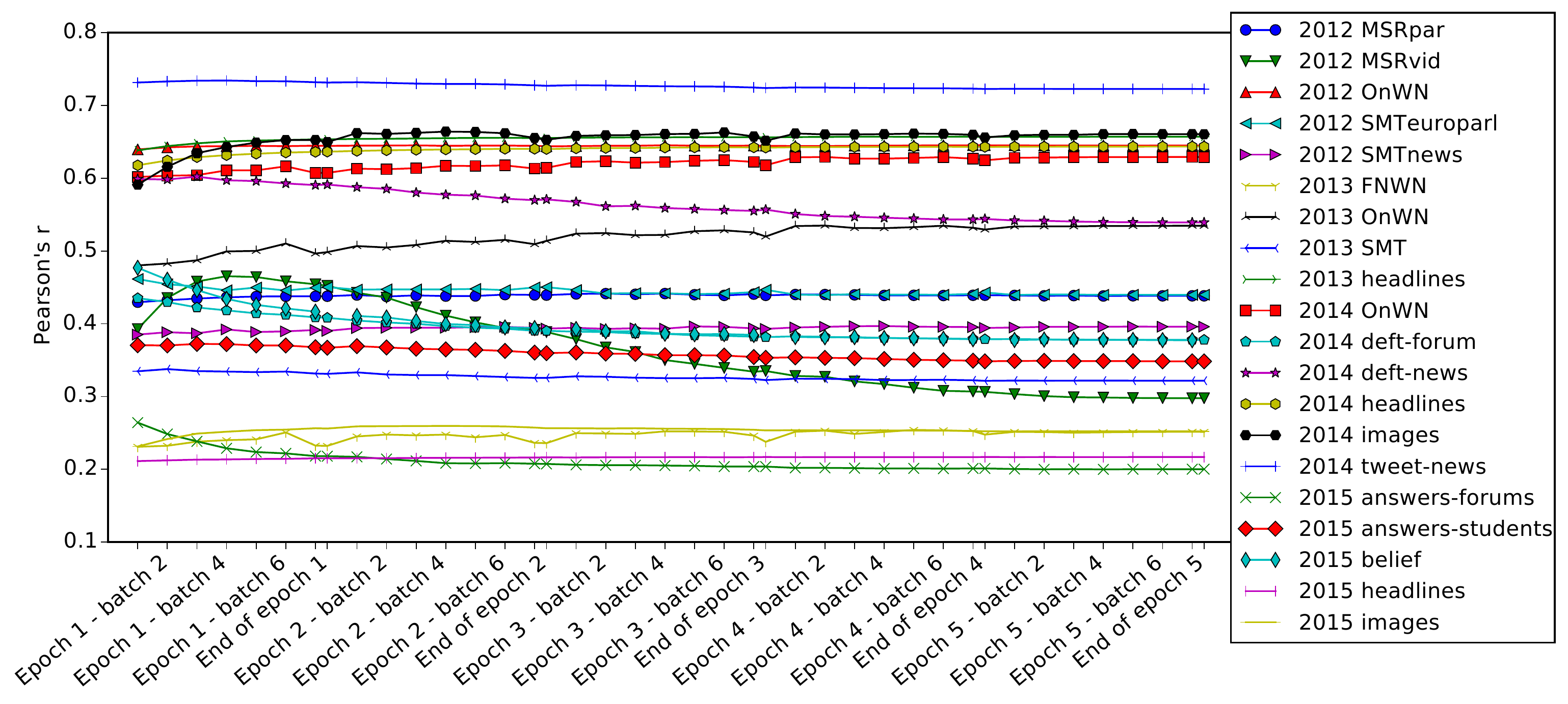}
  \vspace*{-1.4\baselineskip}
  \caption{Performance of Siamese CBOW across 5 iterations.}
  \label{figure:performanceOverTime}
\end{figure*}

Ideally, the optimization criterion of a learning algorithm ranges over the full domain of its loss function.
As discussed in \S\ref{sec:our_approach}, our loss function only observes a sample.
As such, convergence is not guaranteed.
Regardless, an ideal learning system should not fluctuate in terms of performance relative to the amount of training data it observes, provided this amount is substantial: as training proceeds the performance should stabilize.

To see whether the performance of Siamese CBOW fluctuates during training we monitor it during 5 epochs; at every 10,000,000 examples, and at the end of every epoch.
Figure~\ref{figure:performanceOverTime} displays the results for all 20 datasets. We observe that on the majority of datasets the performance shows very little variation.
There are three exceptions.
The performance on the 2014 deft-news dataset steadily decreases while the performance on 2013 OnWN steadily increases, though both seem to stabilize at the end of epoch 5.
The most notable exception, however, is 2012 MSRvid, where the score, after an initial increase, drops consistently.
This effect might be explained by the fact that this evaluation set primarily consists of very short sentences---it has the lowest average sentence length of all set: 6.63 with a standard deviation of 1.812.
Therefore, a 300-dimensional representation appears too large for this dataset; this hypothesis is supported by the fact that  200-dimensional embeddings work slightly better for this dataset (see Figure~\ref{figure:nrOfDimensions}).

\subsubsection{Number of negative examples}\label{sec:nrNegExamples}
\begin{figure}[h]
  \centering
  \includegraphics[width=.49\textwidth]{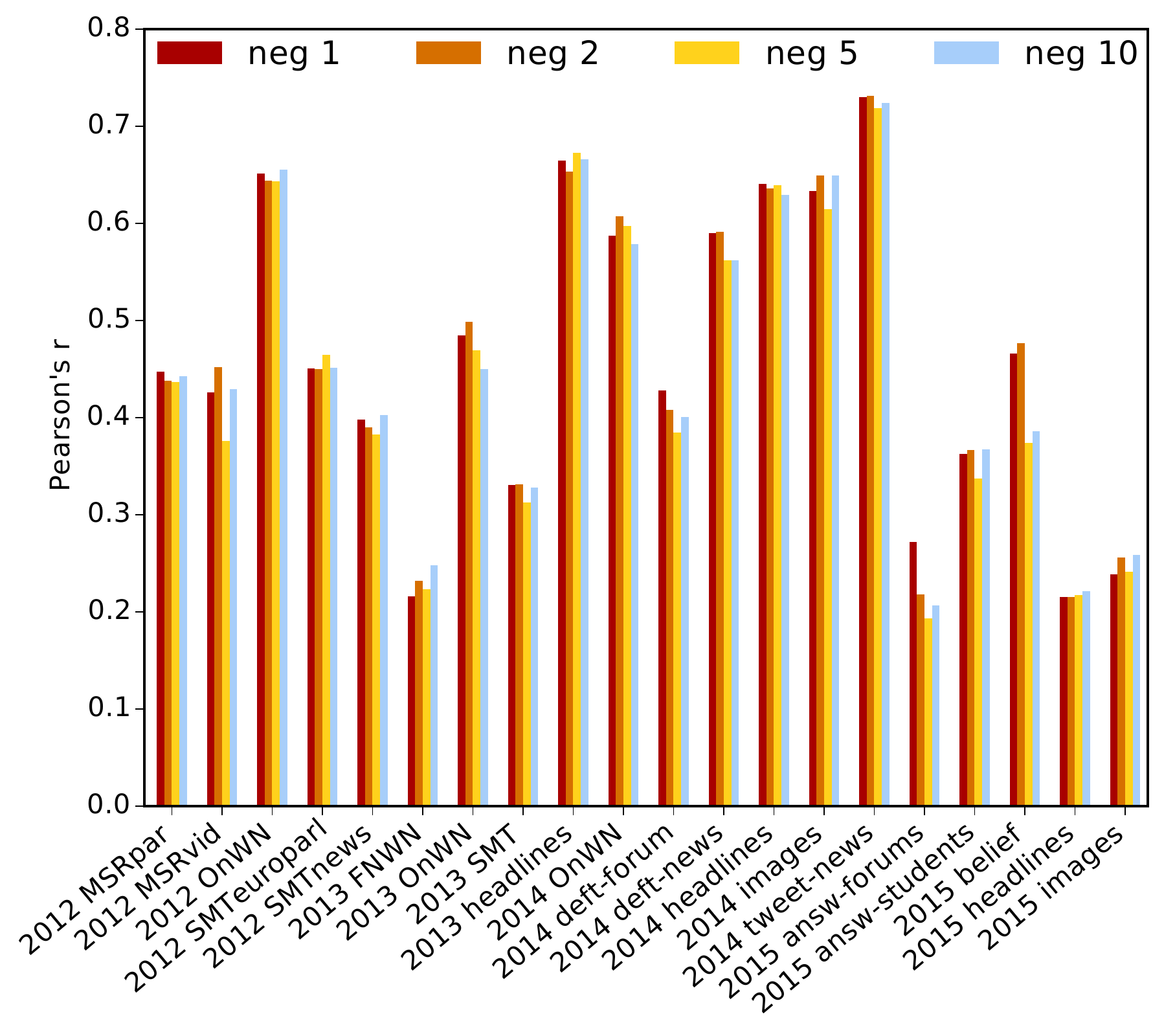}
  \vspace*{-1.4\baselineskip}
  \caption{Performance of Siamese CBOW with different numbers of negative examples.}
  \label{figure:negativeExamples}
\end{figure}

In Figure~\ref{figure:negativeExamples}, the results of Siamese CBOW in terms of Pearson's $r$ are plotted for different numbers of negative examples.
We observe that on most sets, the number of negative examples has limited effect on the performance of Siamese CBOW.
Choosing a higher number, like 10, occasionally leads to slightly better performance, e.g., on the 2013 FNWN set.
However, a small number like 1 or 2 typically suffices, and is sometimes markedly better, e.g., in the case of the 2015 belief set. 
As a high number of negative examples comes at a substantial computational cost, we conclude from the findings presented here that, although Siamese CBOW is robust against different settings of this parameter, setting the number of negative examples to 1 or 2 should be the default choice.


\subsubsection{Number of dimensions}

\begin{figure}[h]
  \centering
  \includegraphics[width=.49\textwidth]{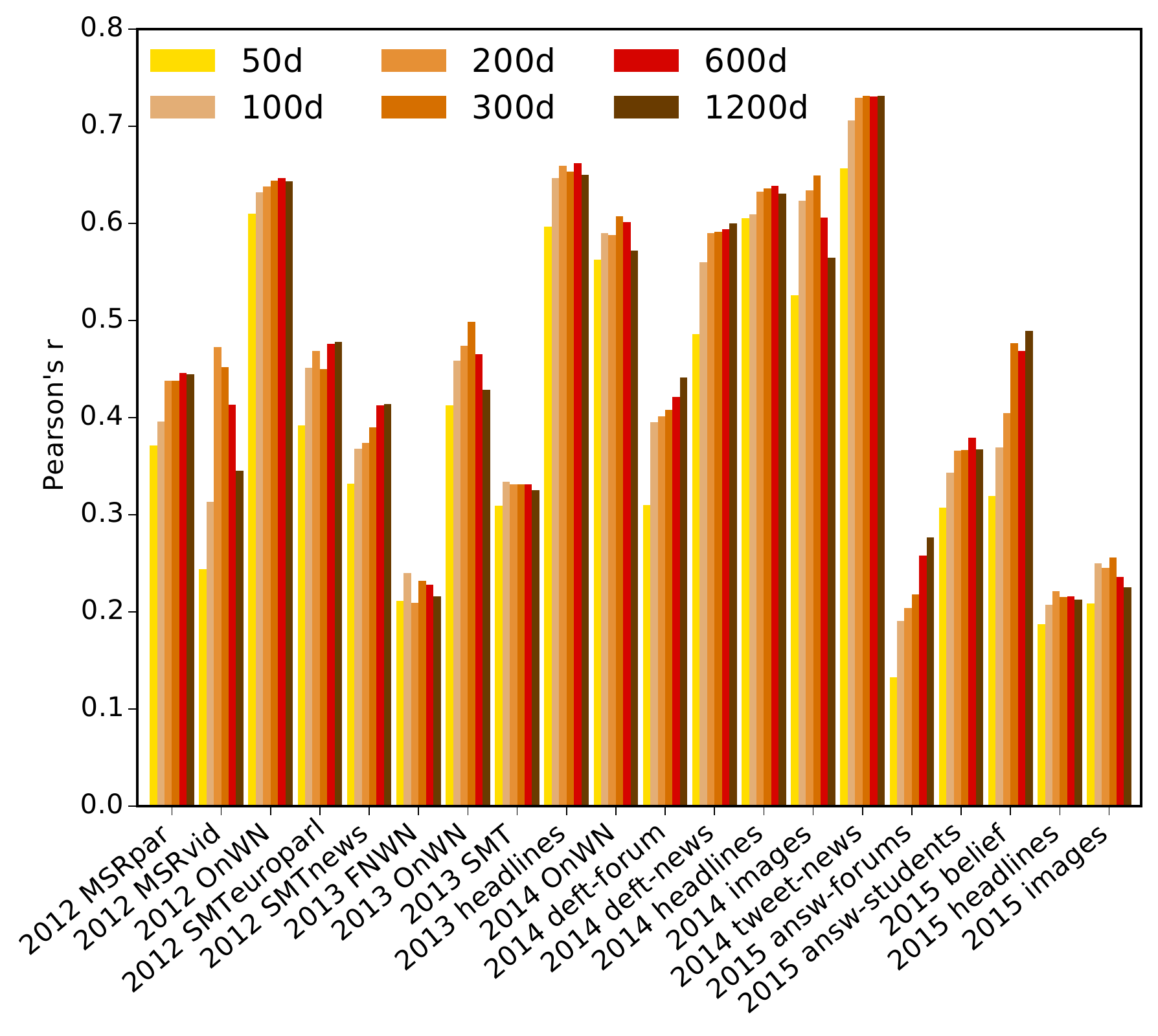}
  \vspace*{-1.4\baselineskip}
  \caption{Performance of Siamese CBOW across number of embedding dimensions.}
  \label{figure:nrOfDimensions}
\end{figure}

Figure~\ref{figure:nrOfDimensions} plots the results of Siamese CBOW for different numbers of vector dimensions.
We observe from the figure that for some sets (most notably 2014 deft-forum, 2015 answ-forums and 2015 belief) increasing the number of embedding dimensions consistently yields higher performance.
A dimensionality that is too low (50 or 100) invariably leads to inferior results.
As, similar to a higher number of negative examples, a higher embedding dimension leads to higher computational costs, we conclude from these findings that a moderate number of dimensions (200 or 300) is to be preferred.

\subsection{Time complexity}\label{sec:efficiency}

For learning systems, time complexity comes into play in the training phase and in the prediction phase.
For an end system employing sentence embeddings, the complexity at prediction time is the most crucial factor, which is why we omit an analysis of training complexity.
We focus on comparing the time complexity for generating sentence embeddings for Siamese CBOW, and compare it to the baselines we use. 

The complexity of all algorithms we consider is $\mathcal{O}(n)$, i.e., linear in the number of input terms.
As in practice the number of arithmetic operations is the critical factor in determining computing time, we will now focus on these.

Both word2vec and the Siamese CBOW compute embeddings of a text $T=t_1, \ldots, t_{|T|}$ by averaging the term embeddings.
This requires $|T| - 1$ vector additions, and 1 multiplication by a scalar value (namely, $\text{1} / |T|$).
The skip-thought model is a recurrent neural network with GRU cells, which computes a set of equations for every term $t$ in $T$, which we reprint for reference~\cite{kiros2015skipthought}:

\vspace*{-1\baselineskip}
{\small
  \begin{eqnarray*}
    \mathbf{r}^t &=& \sigma(\mathbf{W}_{r} \mathbf{x}^t + \mathbf{U}_{r} \mathbf{h}^{t-1})
    \\
    \mathbf{z}^t &=& \sigma(\mathbf{W}_{z} \mathbf{x}^t + \mathbf{U}_{z} \mathbf{h}^{t-1} )
    \\
    \mathbf{\overline{h}}^t &=& \tanh(\mathbf{W}\mathbf{x}^t + \mathbf{U}(\mathbf{r}^t \odot \mathbf{h}^{t-1}))
    \\
    \mathbf{h}^t &=& (1 - \mathbf{z}^t) \odot \mathbf{h}^{t-1} + \mathbf{z}^t \odot \overline{\mathbf{h}}^t
\end{eqnarray*}}

\vspace*{-1.6\baselineskip}\noindent%
As we can see from the formulas, there are $5 |T|$ vector additions (+/-), $4 |T|$ element-wise multiplications by a vector, $3 |T|$ element-wise operations and $6 |T|$ matrix multiplications, of which the latter, the matrix multiplications, are most expensive.

\begin{table}[b]
  \centering
  \caption{Time spent per method on all 20 SemEval datasets, 17,608 sentence pairs, and the average time spent on a single sentence pair (time in seconds unless indicated otherwise).}
  \begin{tabular}{ l c c }
  \toprule
   & 20 sets & 1 pair \\
  \midrule
  Siamese CBOW (300d)  & \phantom{00,00}7.7 & 0.0004 \\
  word2vec (300d)      & \phantom{00,00}7.0 & 0.0004 \\
  skip-thought (1200d) & 98,804.0 & 5.6\phantom{000} \\
  \bottomrule
 \end{tabular}
 \label{table:resultsEfficiency}
\end{table}

This considerable difference in numbers of arithmetic operations is also observed in practice. 
We run tests on a single CPU, using identical code for extracting sentences from the evaluation sets, for every method.
The sentence pairs are presented one by one to the models.
We disregard the time it takes to load models.
Speedups might of course be gained for all methods by presenting the sentences in batches to the models, by computing sentence representations in parallel and by running code on a GPU.
However, as we are interested in the differences between the systems, we run the most simple and straightforward scenario.
Table~\ref{table:resultsEfficiency} lists the number of seconds each method takes to generate and compare sentence embeddings for an input sentence pair.
The difference between word2vec and Siamese CBOW is because of a different implementation of word lookup.

We conclude from the observations presented here, together with the results in \S\ref{sec:main experiments}, that in a setting where speed at prediction time is pivotal, simple averaging methods like word2vec or Siamese CBOW are to be preferred over more involved methods like skip-thought.

\subsection{Qualitative analysis}

As Siamese CBOW directly averages word embeddings for sentences, we expect it to learn that words with little semantic impact have a low vector norm.
Indeed, we find that the 10 words with lowest vector norm are \emph{to}, \emph{of}, \emph{and}, \emph{the}, \emph{a}, \emph{in}, \emph{that}, \emph{with}, \emph{on}, and \emph{as}.
At the other side of the spectrum we find many personal pronouns: \emph{had}, \emph{they}, \emph{we}, \emph{me}, \emph{my}, \emph{he}, \emph{her}, \emph{you}, \emph{she}, \emph{I}, which is natural given that the corpus on which we train consists of fiction, which typically contains dialogues.

It is interesting to see what the differences in related words are between Siamese CBOW and word2vec when trained on the same corpus.
For example, for a cosine similarity $> 0.6$, the words related to \emph{her} in word2vec space are \emph{she}, \emph{his}, \emph{my} and \emph{hers}.
For Siamese CBOW, the only closely related word is \emph{she}.
Similarly, for the word \emph{me}, word2vec finds \emph{him} as most closely related word, while Siamese CBOW comes up with \emph{I} and \emph{my}.
It seems from these few examples that Siamese CBOW learns to be very strict in choosing which words to relate to each other.

\medskip\noindent%
From the results presented in this section we conclude that optimizing word embeddings for the task of being averaged across sentences with Siamese CBOW leads to embeddings that are effective in a large variety of settings. Furthermore, Siamese CBOW is robust to different parameter settings and its performance is stable across iterations. Lastly, we show that Siamese CBOW is fast and efficient in computing sentence embeddings at prediction time.


\section{Related Work}\label{sec:related work}

A distinction can be made between supervised approaches for obtaining representations of short texts, where a model is optimised for a specific scenario, given a labeled training set, and unsupervised methods, trained on unlabeled data, that aim to capture short text semantics that are robust across tasks.
In the first setting, word vectors are typically used as features or network initialisations \cite{kenter-short-2015,hu2014convolutional,severyn2015learningtr,yin2015convolutional}.
Our work can be classified in the latter category of unsupervised approaches.

Many models related to the one we present here are used in a multilingual setting \cite{hermann2014multilingual,hermann2014iclr,lauly2014autoencoder}.
The key difference between this work and ours is that in a multilingual setting the goal is to predict, from a distributed representation of an input sentence, the same sentence in a different language, whereas our goals is to predict surrounding sentences.

\newcite{Wieting2015TowardsUP} apply a model similar to ours in a related but different setting where explicit semantic knowledge is leveraged.
As in our setting, word embeddings are trained by averaging them. 
However, unlike in our proposal, a margin-based loss function is used, which involves a parameter that has to be tuned.
Furthermore, to select negative examples, at every training step, a computationally expensive comparison is made between all sentences in the training batch.
The most crucial difference is that a large set of phrase pairs explicitly marked for semantic similarity has to be available as training material.
Obtaining such high-quality training material is non-trivial, expensive and limits an approach to settings for which such material is available.
In our work, we leverage unlabeled training data, of which there is a virtually unlimited amount. 

As detailed in \S\ref{sec:our_approach}, our network predicts a sentence from its neighbouring sentences.
The notion of learning from context sentences is also applied in \cite{kiros2015skipthought}, where a recurrent neural network is employed.
Our way of averaging the vectors of words contained in a sentence is more similar to the CBOW architecture of word2vec \cite{mikolov_efficient_2013}, in which all context word vectors are aggregated to predict the one omitted word.
A crucial difference between our approach and the word2vec CBOW approach is that we compare sentence representations directly, rather than comparing a (partial) sentence representation to a word representation.
Given the correspondence between word2vec's CBOW model and ours, we included it as a baseline in our experiments in \S\ref{sec:experimental setup}.
As the skip-gram architecture has proven to be a strong baseline too in many settings, we include it too.

\newcite{yih2011learningdp} also propose a siamese architecture.
Short texts are represented by tf-idf vectors and a linear combination of input weights is learnt by a two-layer fully connected network, which is used to represent the input text.
The cosine similarity between pairs of representations is computed, but unlike our proposal, the differences between similarities of a positive and negative sentence pair are combined in a logistic loss function.

Finally, independently from our work, \newcite{hill2016learning} also present a log-linear model.
Rather than comparing sentence representations to each other, as we propose, words in one sentence are compared to the representation of another sentence.
As both input and output vectors are learnt, while we tie the parameters across the entire model, \newcite{hill2016learning}'s model has twice as many parameters as ours.
Most importantly, however, the cost function used in \cite{hill2016learning} is crucially different from ours.
As words in surrounding sentences are being compared to a sentence representation, the final layer of their network produces a softmax over the entire vocabulary.
This is fundamentally different from the final softmax over cosines between sentence representations that we propose.
Furthermore, the softmax over the vocabulary is, obviously, of vocabulary size, and hence grows when bigger vocabularies are used, causing additional computational cost.
In our case, the size of the softmax is the number of positive plus negative examples (see \S\ref{sec:our_approach:training_objective}).
When the vocabulary grows, this size is unaffected. 


\section{Conclusion}\label{sec:conclusion}

We have presented Siamese CBOW, a neural network architecture that efficiently learns word embeddings optimized for producing sentence representations.
The model is trained using only unlabeled text data.
It predicts, from an input sentence representation, the preceding and following sentence.

We evaluated the model on 20 test sets and show that in a majority of cases, 14 out of 20, Siamese CBOW outperforms a word2vec baseline and a baseline based on the recently proposed skip-thought architecture.    
As further analysis on various choices of parameters show that the method is stable across settings, we conclude that Siamese CBOW provides a robust way of generating high-quality sentence representations.



Word and sentence embeddings are ubiquitous and many different ways of using them in supervised tasks have been proposed.
It is beyond the scope of this paper to provide a comprehensive analysis of all supervised methods using word or sentence embeddings and the effect Siamese CBOW would have on them.
However, it would be interesting to see how Siamese CBOW embeddings would affect results in supervised tasks.

Lastly, although we evaluated Siamese CBOW on sentence pairs, there is no theoretical limitation restricting it to sentences.
It would be interesting to see how embeddings for larger pieces of texts, such as documents, would perform in document clustering or filtering  tasks.


\section*{Acknowledgments}

The authors wish to express their gratitude for the valuable advice and relevant pointers of the anonymous reviewers.
Many thanks to Christophe Van Gysel for implementation-related help.
This research was supported by
Ahold,
Amsterdam Data Science,
the Bloomberg Research Grant program,
the Dutch national program COMMIT,
Elsevier,
the European Community's Seventh Framework Programme (FP7/2007-2013) under
grant agreement nr 312827 (VOX-Pol),
the ESF Research Network Program ELIAS,
the Royal Dutch Academy of Sciences (KNAW) under the Elite Network Shifts project,
the Microsoft Research Ph.D.\ program,
the Netherlands eScience Center under project number 027.012.105,
the Netherlands Institute for Sound and Vision,
the Netherlands Organisation for Scientific Research (NWO)
under pro\-ject nrs
727.\-011.\-005, 
612.001.116, 
HOR-11-10, 
640.006.013, 
612.\-066.\-930, 
CI-14-25, 
SH-322-15, 
652.\-002.\-001, 
612.\-001.\-551, 
%
the Yahoo Faculty Research and Engagement Program,
and
Yandex.
All content represents the opinion of the authors, which is not necessarily shared or endorsed by their respective employers and/or sponsors.

\bibliography{acl2016-fp-ssn}
\bibliographystyle{acl2016}

\end{document}